\documentclass[journal]{IEEEtran}
\ifCLASSINFOpdf
\else
\fi

\usepackage{balance}
\usepackage{graphicx}
\usepackage{color}
\usepackage[ruled, linesnumbered]{algorithm2e}
\usepackage{subcaption}
\usepackage{amsmath}
\usepackage{amssymb}
\usepackage{tikz}
\usepackage{etoolbox}

\graphicspath{{imgs/}}

\begin{document}
%
\title{Tool Macgyvering: A Novel Framework for Combining Tool Substitution and Construction}
%
%
%

\author{Lakshmi Nair, Nithin Shrivatsav and Sonia Chernova

\thanks{All authors are affiliated with the College of Computing at Georgia Institute of Technology, Atlanta, GA, USA (Corresponding author: lnair3@gatech.edu)}}

%
%

\markboth{Submitted to IEEE Transactions on Robotics, August~2020}%
{Nair \MakeLowercase{\textit{et al.}}: Tool Macgyvering}
%



\maketitle

\begin{abstract}
Macgyvering refers to solving problems inventively by using whatever objects are available at hand. Tool Macgyvering is a subset of macgyvering tasks involving a missing tool that is either substituted (tool substitution) or constructed (tool construction), from available objects. In this paper, we introduce a novel Tool Macgyvering framework that combines tool substitution and construction using arbitration that decides between the two options to output a final macgyvering solution. Our tool construction approach reasons about the shape, material, and different ways of attaching objects to construct a desired tool. We further develop value functions that enable the robot to effectively arbitrate between substitution and construction. Our results show that our tool construction approach is able to successfully construct working tools with an accuracy of 96.67\%, and our arbitration strategy successfully chooses between substitution and construction with an accuracy of 83.33\%.
\end{abstract}


%
\IEEEpeerreviewmaketitle

\section{Introduction}

A transformative change for robotics is enabling robots to effectively improvise tools. Tools can extend the physical capabilities of robots and make them more useful, by enabling them to go beyond small, fixed sets of interchangeable end-effectors often found in industrial settings. However, a major problem with the philosophy of emphasizing tool use is that the right tool is not always accessible, and robots may have to improvise with what is available. Humans, chimpanzees and certain species of birds have all been known to accomplish tasks by creatively utilizing objects available to them, such as sticks and stones \cite{stout2011stone, toth1993pan, jones1973tool}. In the Apollo 13 incident of 1970, a carbon dioxide filter creatively constructed out of a sock, a plastic bag, book covers, and duct tape helped save the lives of the three astronauts on board \cite{cass2005apollo}. Solving problems inventively by using available objects is colloquially referred to as Macgyvering, and is featured extensively in TV shows \cite{Macgyver}, books \cite{Martian}, inventions \cite{RubeGoldberg}, and even cultural traditions (e.g., the Indian tradition of Jugaad \cite{Jugaad}). However, similar tool improvisation and macgyvering capabilities are currently beyond the scope of robots today, limiting them to predefined tools and tasks. The ability to improvise and invent appropriate tools from available resources can greatly increase robot adaptability, enabling robots to handle any uncertainties or equipment failures that may arise. These capabilities will be particularly useful for robots that explore, as well as work in space, underwater, and other locations where required tools may not be easily available. 

\begin{figure}[t]
	\centering
	\includegraphics[width=0.48\textwidth]{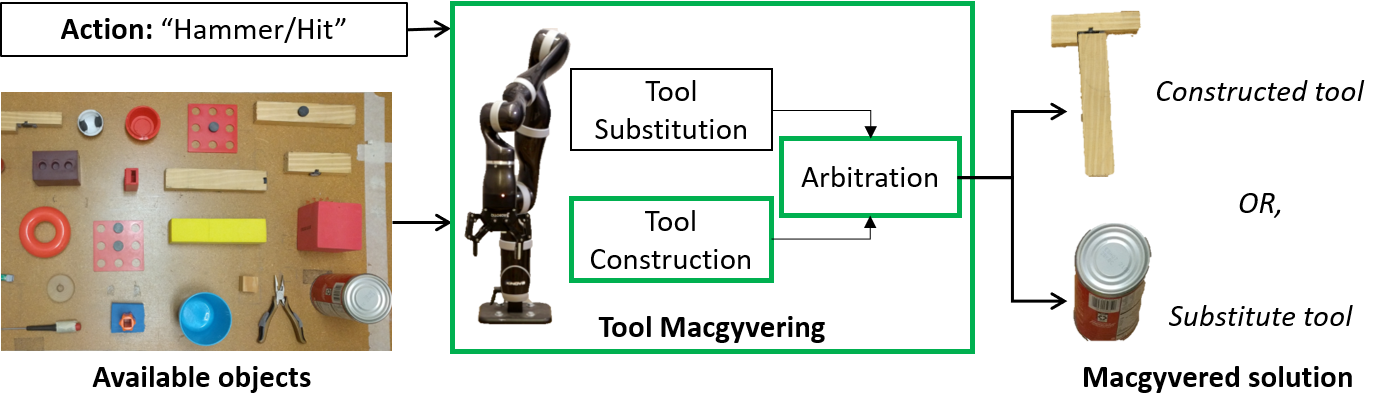}
	\captionsetup{width=\linewidth}
	\caption{Tool macvgyering: Given an action or task, and available objects, the robot either substitutes for the missing tool (e.g., using a metal can), or constructs a tool for performing the action (e.g., making a hammer by joining two objects). Highlighted in green are the key contributions of this paper.}
	\label{fig:algo_overview}
\end{figure}

Our goal in this work is to enhance the adaptability of robots beyond predefined or prototypical tools. We seek to enable robots to improvise or \textit{Macgyver} tool-based solutions, either by directly selecting an available substitute for a missing tool through tool substitution, or building an appropriate tool from available objects through tool construction. Specifically, we define \textit{Tool Macgyvering} as a subset of macgyvering problems involving tool substitution (e.g., a metal can is used as a substitute tool for hammering a nail), or tool construction (e.g., a hammer is constructed from wooden pieces).


In this paper, we contribute a novel Tool Macgyvering framework that takes in a set of available objects along with a desired action to be performed (e.g., ``hit/hammer''), and outputs either a tool substitute or tool construction for performing the action. An overview of our Tool Macgyvering framework is shown in Figure \ref{fig:algo_overview}. \textbf{Tool substitution} identifies the most appropriate object for performing the action, by reasoning about the shape and material of the available objects. \textbf{Tool construction} identifies the most appropriate object combination (construction) for performing the action, by reasoning about shape, material and the different ways of attaching the objects. Finally, \textbf{arbitration} selects between the object substitutes and constructions to output the most appropriate Tool Macgyvering solution.
 

For performing tool substitution, we directly apply our prior work that has been shown to effectively identify substitute tools from partial point clouds \cite{shrivatsav2020}. For tool construction, our prior work introduced an approach that only reasoned about shape and potential ways of attaching objects together to construct tools \cite{nair2019toolconstr, nair2019autonomous}. However, this resulted in tools made of inappropriate materials, e.g., hammers made out of foam. In this paper, we extend our tool construction approach by incorporating material reasoning, to enable robots to reason about material properties when constructing tools. We further evaluate our current approach against our prior work, with an expanded test set of objects of varied shapes and materials, allowing for the construction of more diverse tools than before. As we show in our experiments, the presented framework results in significant improvement over prior results, both in terms of performance and the quality of output constructions. 

Our key contributions in this paper are as follows:
\begin{enumerate}
    \item Introduction of a novel, unified Tool Macgyvering framework that combines tool substitution and construction, using arbitration to decide between substitution and construction as the appropriate macgyvering solution;
    \item Incorporation of material reasoning for tool construction, that significantly improves performance over previous tool construction approaches; 
    \item Introduction of arbitration strategies for selecting between tool substitution and tool construction.
\end{enumerate}

We validate the effectiveness of our tool construction approach on a 7-DOF robotic arm, through autonomous construction of six different tool types. We also demonstrate the efficiency of our arbitration approaches in deciding the most appropriate macgyvering solution for a specified action.

\section{Related Work}
In this section, we summarize existing work that is closely related to Tool Macgyvering.



\subsection{Tool Construction}
Existing research in robotics has primarily focused on tool use (\cite{sinapov2007learning, sinapov2008detecting}), with little prior work in tool construction. Some recent work has explored Macgyvering and the inventive use of available objects for problem solving \cite{sarathy2017macgyver, sarathy2018macgyver}. They propose a theoretical formulation of Macgyvering problems as scenarios that require the initial domain to be transformed (e.g., by adding a state or action), for the goal state to be reachable. They further introduce the Macgyver Test as an alternative to the Turing test, to measure the resourcefulness of robots. Our work differs from theirs in that we explicitly reason about visual and physical properties of objects, and different ways of attaching objects for Tool Macgyvering. 

Additional research in macgyvering has also focused on the construction of environmental structures, such as techniques for Automated Design of Functional Structures (ADFS), involving construction of navigational structures, e.g., stairs or bridges \cite{erdogan2013planning}. They introduce a framework for effectively partitioning the solution space by inducing constraints on the design of the structures. Further, \cite{tosun2018perception} has looked at planning for construction of functional structures by modular robots, focusing on identifying features that enable environmental modification in order to make it traversable. In similar work, \cite{saboia2018autonomous} has looked at modification of unstructured environments using objects, to create ramps that enhance navigability. More recently, \cite{choi2018creating} extended the cognitive architecture ICARUS to support the creation and use of functional structures such as ramps, in abstract planning scenarios. The formulation of the problem specifically conforms to the cognitive architecture, limiting its generalization. More broadly, these approaches are primarily focused on improving robot navigation through environment modification as opposed to construction of tools. 

Some existing research has also explored the construction of simple machines such as levers and bridges \cite{stilman2014robots, levihn2014using}. Their work formulates the construction of simple machines as a constraint satisfaction problem where the constraints represent the relationships between the design components. The constraints in their work limit the variability of the simple machines that can be constructed, focusing only on the placement of components relative to one another, e.g., placing a plank over a stone to create a lever. Additionally, \cite{wicaksono17towards} focused on using 3D printing to fabricate tools from polymers. However, these approaches do not address the problem of tool construction using environmental objects. 


\subsection{Tool Substitution}
Prior work in tool substitution has explored the use of large-scale semantic networks \cite{boteanu2015towards}, or visual similarities between tools (\cite{abelha2016model, schoeler2016bootstrapping}), to identify good substitutes. In \cite{abelha2016model}, the authors use Superquadrics (SQs) to model objects for tool substitution. SQs are geometric shapes that include quadrics, but allows for arbitrary powers instead of just power of two. In their approach, the candidate tools are represented using SQ parameters, and compared to the desired parameters of the tool for which a replacement is sought. In \cite{schoeler2016bootstrapping}, they learn function-to-shape correspondence of objects using supervised learning to identify substitutes for a given tool using part-based shape matching. To model the tools, they use existing point cloud shape representations, such as Ensemble of Shape Functions (ESF) \cite{wohlkinger2011ensemble}. ESF is a descriptor consisting of 10, 64-bin sized histograms (640-D vector), describing the shape of a point cloud, with much success in representing partial point clouds \cite{wohlkinger2011ensemble, nair2019autonomous}. However, these approaches do not reason about material of the objects when evaluating the substitutes. 


\subsection{Arbitration of Behaviors}
While there has not been prior work specifically in arbitration of tool substitution and construction, behavior-based design methodologies often explore coordination mechanisms for different robot behaviors \cite{arkin2003ethological, likhachev2000robotic, velayudhan2017sloth}. Given a set of behaviors, the goal of the coordination or arbitration mechanism is to generate an output behavior that is either one, or a combination of the input behaviors. Two arbitration strategies have been commonly explored to accomplish this, namely, Action Selection and Behavioral Fusion \cite{velayudhan2017sloth}. In action selection, each behavior is associated with a value function that dictates the behavior chosen at any given instant. Thus, only one of the input behaviors is selected. In contrast, behavioral fusion generates a weighted summation of the input behaviors, often used in navigational tasks. Given the nature of our problem, we use action selection to arbitrate between tool substitution and construction, developing appropriate value functions to select the desired behavior for the specified action. 

\subsection{Material Reasoning}
Material properties play an important role when detecting appropriate objects for tool construction and substitution, e.g., for hammering, wooden or metallic objects are preferred over foam. In \cite{perlow2017raw}, the authors describe an approach for detecting appropriate raw materials for object construction, and demonstrate their work in the simulated world of Minecraft. Their work uses neural networks to classify materials from object images. Several other vision-based approaches to material recognition have been previously explored \cite{schwartz2018recognizing, hu2011toward, bell2015material}. These approaches focus on the visual appearance of objects to decipher their material properties. In contrast to visual reasoning, \cite{erickson2019classification} has explored the use of spectral reasoning for material classification. Spectral reasoning uses a handheld spectrometer to measure the reflected intensities of different wavelengths, in order to profile and classify object materials. Their work has shown promising results with a validation accuracy of 94.6\%. However, generalizing posed a greater challenge, achieving an accuracy of 79\% on previously unseen objects. Nevertheless, spectral data helps offset some critical deficiencies of vision-based approaches, such as sensitivity to light and viewing angle. In this work, we focus on using spectral data to reason about materials of the objects.

\subsection{Dual Neural Networks}
Dual neural networks\footnote{Also known as Siamese neural networks. We avoid using the term ``Siamese'', instead referring to such networks as Dual Neural Networks in our paper} consist of two identical networks, each accepting a different input, combined at the end with a distance metric. The parameters of the twin networks are tied, and the distance metric computes difference between the final layers of the twin networks. Prior work has successfully used dual networks for matching images \cite{koch2015siamese, bromley1994signature, schroff2015facenet}. In this work, we use dual networks to perform shape and material scoring for tool construction and substitution. Our approach is similar to FaceNet \cite{schroff2015facenet}, in that we learn an embedding from the training data, which is then used to match a query input by computing a similarity score. However, in contrast to prior work, the inputs to our dual networks use ESF features for shape scoring, and spectral data for material scoring.

\begin{figure*}[t]
    \centering
    \includegraphics[width=1.0\textwidth]{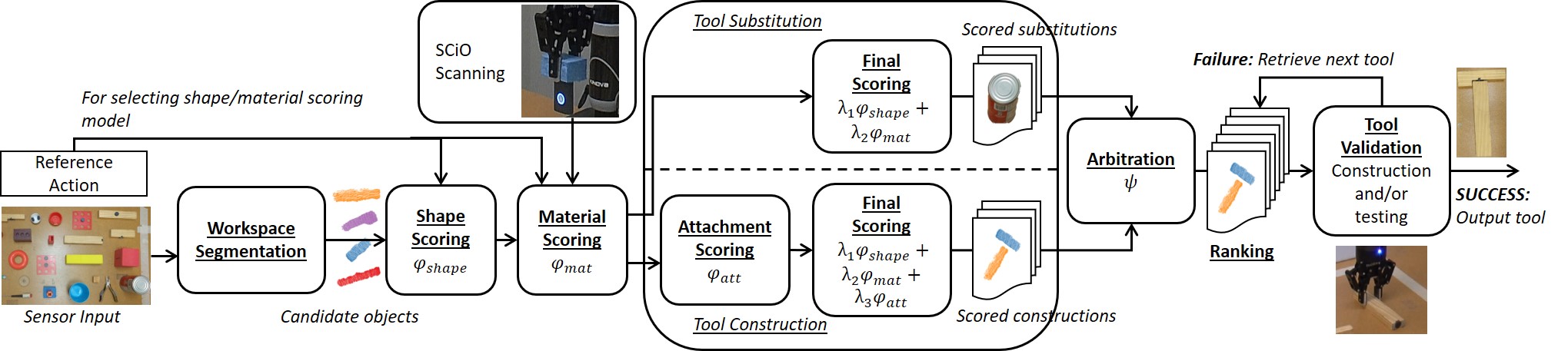}
    \captionsetup{width=\linewidth}
    \caption{Overview of our Tool Macgyvering framework highlighting the different steps involved. The tool construction and substitution pipelines are followed by arbitration, to output a combined ranking of the different strategies that the robot then validates. Arbitration essentially combines substitution and construction within the framework.}
    \label{fig:framework}
\end{figure*}

\section{Tool Macgyvering}
In this section we introduce our Tool Macgyvering framework. We begin by formulating our primary research problem as follows: 

\textit{``Given an action, and a set $C$ of $n$ candidate objects, how can we generate an output ranking of macgyvered solutions for accomplishing the specified action?''}
\smallskip


We denote the set of all candidate objects as $C = \{c_1, c_2, ..., c_n\}$. Thus, the problem of identifying tool substitutes involves a search space of size $n$, where each $c_i$ is a potential substitute. However, tool construction presents a more challenging combinatorial state space of size $^nP_m$, assuming that we wish to construct a tool with $m$ objects. We denote the set of all permutations of the $m$ objects as $T = \{T_1, T_2, ...\}$, where $T_i = (c_1, ..., c_m)$ is a tuple representing a specific permutation of $m$ objects. We denote the combined space of tool substitutions and constructions as, $S = C \cup T$, where $|S| = n + {^nP_m}$. The goal of our approach is to evaluate the states in $S$, to identify the best Tool Macgyvering solution.

In order to identify the most suitable set of objects, we develop objective functions that effectively score the appropriateness of the substitutes and constructions for performing the specified action. A general objective function can be expressed as a weighted sum over a set of $k$ features of the candidate objects in $s_i \in S$, denoted by $\phi_1, ..., \phi_k$, as follows:
\[\Phi(s_i) = \lambda_1*\phi_1(s_i) + \lambda_2*\phi_2(s_i) + ... + \lambda_k*\phi_k(s_i)\]
We show that reasoning about three features of the candidate objects, namely, \textbf{\textit{shape}} ($\phi_{shape}$), \textbf{\textit{materials}} ($\phi_{mat}$), and \textbf{\textit{attachments}} ($\phi_{att}$) in the case of tool constructions, enables the robot to effectively explore the state space. We define \textit{attachments} as locations at which objects can be attached together. Our work introduces a learning-based framework for computing the objective function, that is computationally scalable as number of objects increases. 

\begin{algorithm}[t]
		\SetKwInOut{Input}{input}\SetKwInOut{Output}{output}
		
		\Input{action; $T=permute(C,m)$}
		\Output{$T^*$, $Att$, $Type$}
		\BlankLine
		
		$E = [], Att = [], Type = []$
		
		$S = C \cup T$
		
		\For{$i\gets1$ \KwTo $|S|$}{
			
			$\phi_{shape}(s_i) = ShapeFit(s_i, action)$ 
			
			$\phi_{mat}(s_i) = MaterialFit(s_i, action)$
			
			\uIf{$|s_i| > 1$}{ \tcp{Construction with $T_i$}
			
			    $t_{att} = AttachType(s_i)$
			
			    $\phi_{att}(s_i), A_{close}(s_i) = AttachmentFit(s_i, t_{att})$
			    
			    $\Phi(s_i) = \phi_{shape}(s_i) + \phi_{mat}(s_i) + \phi_{att}(s_i)$
			    
			}
			\Else{ \tcp{Substitution with $c_i$} 
			    
			    $t_{att} = \varnothing$
			    
			    $A_{close}(s_i) = \varnothing$
			    
			    $\Phi(s_i) = \phi_{shape}(s_i) + \phi_{mat}(s_i)$
			    
			}
			
			$E.append(\Phi(s_i))$
			
			$Att.append(A_{close}(s_i))$
			
			$Type.append(t_{att})$
			
		}
		
		\tcp{Arbitrate based on value functions}
		
		$V = Arbitrate(E, S)$ 
		
		$S^* = sort(S, V)$ \tcp{Sort $S$ based on $V$}
		
		\Return $S^*, Att, Type$
		
		\caption{Tool Macgyvering}
\end{algorithm}

Our complete Tool Macgyvering framework is shown in Figure \ref{fig:framework}, and our complete Tool Macgyvering algorithm is shown in Algorithm 1. The pipeline begins with \textbf{workspace segmentation} which enables the system to identify the candidate objects in the robot's workspace. We use plane subtraction and Sample Consensus Segmentation (SAC)\footnote{The implementation was provided by the PCL library} to identify the candidate objects available to the robot using RGB-D data from a camera mounted over the table. The \textbf{shape scoring} algorithm ($ShapeFit()$, Algorithm 1, line 4), evaluates the visual appropriateness of the candidate objects and assigns a corresponding shape score ($\phi_{shape}$ or $\phi'_{shape}$). In this paper, we present two ways of computing the shape score, detailed in the following sections. Following shape scoring, the \textbf{material scoring} algorithm ($MaterialFit()$, Algorithm 1, line 5), evaluates the material fitness of the candidate objects, and assigns a corresponding material score ($\phi_{mat}$). The shape and material scores are combined for tool substitution, in a final objective function $\Phi^{subs}$. For tool construction, the scores discussed above do not indicate whether the objects can be attached. Hence, our \textbf{attachment scoring} algorithm ($AttachmentFit()$, Algorithm 2), evaluates whether the candidate objects can be attached. The algorithm outputs an attachment score, which is combined with the shape and material scores to compute the final objective function $\Phi^{cons}$. The final objectives are then used for computing value functions for \textbf{arbitration}. Arbitration uses the value functions to generate a combined \textbf{ranking} of the tool substitutes and constructions (ranked from highest to lowest values). Finally, the robot \textbf{validates} each construction/substitute for their task suitability, by applying the desired action with the object. In the case of construction, the robot first constructs the tool, and then validates it by applying the desired action on the tool. In this work, we assume that the robot can observe whether the tool succeeded, and that the action trajectory is pre-specified. Alternatively, the action trajectory could be learned from demonstration \cite{rana2017towards}, including, if necessary, adapting the original action to fit the dimensions of the new tool \cite{fitzgerald2014representing,gajewski2018adapting}. If the object fails at performing the action or cannot be constructed, the robot iterates through the ranks until a solution is found.

In the following sections we describe material, shape, and attachment scoring, followed by the final objective computation for tool substitution and tool construction. Finally, we present three different arbitration strategies for ranking the substitutes and constructions.


\subsection{Material Scoring ($\phi_{mat}$)}

Given an action and spectral reading of an object as inputs, material scoring seeks to predict the degree to which the spectral reading is similar to that of canonical tools used for the action. Our previous work has shown that supervised learning using dual neural networks is able to effectively predict material similarity between objects (\cite{shrivatsav2020}), and we follow a similar approach to compute $\phi_{mat}$. 

Dual neural networks consist of two identical networks, each accepting a different input, combined at the end with a distance metric. The parameters of the twin networks are tied, and the distance metric computes difference between the final layers of the twin networks. The networks are trained on pairs of inputs that are of the same/different classes, to discriminate between the class identity of the input pairs. Once the network weights are learned, we use positive examples (i.e., canonical materials) from the training data to learn an \textit{embedding}. $\phi_{mat}$ is then computed as the similarity of the query spectral reading to the embedding. This enables us to match the input spectral reading to the variety of canonical materials that facilitate an action, rather than conforming to the materials of a specific tool. Here, we assume that the material of the action part of the tool is most critical to performing the action.  As a result, we simplify our model by only considering the material of the action part, e.g., we model a knife consisting of a metal blade and plastic handle, as metal. This assumption holds for the vast majority of household tools, but could be relaxed in future work.

\begin{table}[t]
	\centering
	\includegraphics[width=0.38\textwidth]{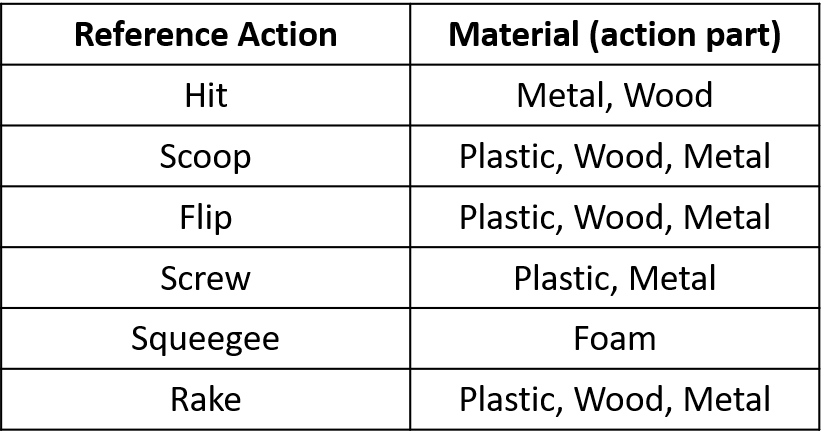}
	\captionsetup{width=\linewidth}
	\caption{Showing the appropriate materials for performing each action, used for generating training pairs for the dual networks}
	\label{tbl:materials}
\end{table}

\subsubsection{Feature Representation}
We use the SCiO, a commercially available handheld spectrometer (shown in Figure \ref{fig:framework}), to extract spectral readings for the objects. The SCiO scans objects to return a 331-D vector of real-valued spectral readings. 

\subsubsection{Network Architecture}
Our model consists of three hidden layers of 426, 284 and 128 units each. We apply tanh activation and a dropout of 0.5 after each layer. The final layer is a sigmoid computation over the element-wise $L_1$ difference between the third layer of the two networks. We use Adam optimizer with learning rate of 0.001.  
 
\subsubsection{Training}
To train the dual neural network, we use the SMM50 dataset\footnote{Dataset available at https://github.com/Healthcare-Robotics/smm50}, which contains spectrometer readings for five classes of materials: plastic, paper, wood, metal and foam. For our work, we manually identified the most appropriate material classes for different actions, also shown in Table \ref{tbl:materials}. We create random pairings of spectral readings, where  both materials in the pair are appropriate for the action, or either one is not. Given a set $N$ of training samples, $y(x_i, x_j) = 1$, if both materials are appropriate for a given action (as indicated by Table \ref{tbl:materials}), and $y(x_i, x_j) = 0$, if either $x_i$ or $x_j$ corresponds to an inappropriate material. That is, for ``Hit'', (metal, metal) and (metal, wood) pairings are both positive examples, whereas (metal, foam) is a negative example. Note that, each pair does not necessarily consist of the same material class. The reason is that, we would like all appropriate material classes for a given action, such as metal and wood for ``Hit'', to be mapped closer in the embedding space, than metal and foam. This allows us to overcome the variance across material classes, learning an embedding space where the desired material classes are closer in distance. Our training minimizes the standard regularized binary cross-entropy loss function as:
\begin{align*}
\mathcal{L}(x_i, x_j) = y(x_i, x_j)\log(\mathbf{p}(x_i, x_j)) + \\ (1-y(x_i,x_j))\log(1-\mathbf{p}(x_i, x_j)) + \lambda |\mathbf{w}|^2
\end{align*}
The output prediction of the final layer $L$, is given as:
\begin{align*}
    \mathbf{p} = \sigma (\mathbf{w}^T (|h_{1, L-1} - h_{2, L-1}|) + \beta)
\end{align*}
Where $\sigma$ denotes the sigmoidal activation function, $\beta$ denotes the bias term learned during training, and $h_{1, L-1}$, $h_{2, L-1}$ denotes the final hidden layers of the twin networks respectively. The element-wise $L_1$ norm of the final hidden layers is passed to the sigmoid function. In essence, the sigmoid function computes a similarity between the output features of the final hidden layers of the two twin networks.

Once the network is trained, we learn an embedding using the positive examples (not pairings) from our training set, $x^p_i \in N$, where $x^p_i$ is an appropriate spectral reading for the action. We denote the output of the final hidden layer, for a given input $x$ as, $f(x) = h_{1, L-1}(x)$. We pass each $x^p_i$ through one of the twin networks (since both networks are identical and their weights tied), to map each input into a $d$-dimensional Euclidean space, denoted by $f(x^p_i) \in \mathbb{R}^d$. We then compute the embedding as an average over $f(x^p_i)$, for all the positive examples $x^p_i$, where $N_p$ is the total number of positive examples in the training set:
\begin{align*}
    \mathcal{D}^p_{action} = \frac{1}{N_p} \sum_{i=1}^{N_p} f(x^p_i) \ \forall \ x^p_i \in N
\end{align*}
We compute the $d$-dimensional embedding space $\mathcal{D}^p_{action}$, using the spectral readings corresponding to appropriate materials as positive examples, $x^p_i \in N$. The computed embedding represents an aggregation of the most appropriate spectral readings in the training set for a specific action. 

\subsubsection{Prediction}
Given the spectral reading corresponding to a candidate object $c_j$, we compute $f(c_j)$ using our pre-trained model. Then, $\phi_{mat}$ is computed by $MaterialFit()$ (Algorithm 1, Line 5), as follows:
\begin{align*}
    \phi_{mat}(c_j) = \sigma (\mathbf{w}^T |\mathcal{D}^p_{action} - f(c_j)| + \beta)
\end{align*}
This score represents the similarity between material of the candidate object and the embedding, $\mathcal{D}^p_{action}$, representative of all the positive examples within the training data. For tool construction, the score is computed for the objects $c_j \in T_i$. 

\begin{figure}[t]
    \centering
    \includegraphics[width=0.49\textwidth]{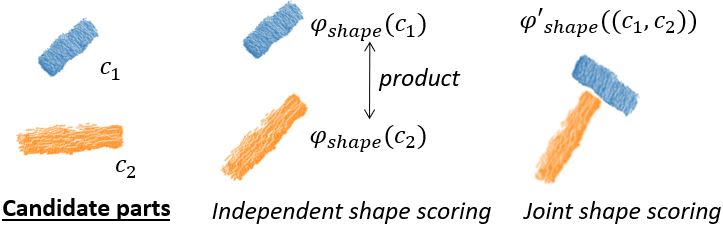}
    \captionsetup{width=\linewidth}
    \caption{Figure highlighting the two types of shape scoring. For independent shape scoring, the candidate parts are scored independently and combined into a single score as a product of their independent scores. For joint shape scoring, the composite object is scored.}
    \label{fig:shape_scoring}
\end{figure}

\subsection{Shape Scoring ($\phi_{shape}$ or $\phi'_{shape}$)}
Given an action, e.g., ``scoop'', and object point cloud as inputs, shape scoring seeks to predict the shape fitness of the object for performing the action, by learning shape-to-function correspondence of objects. In this paper, we present two ways of computing the shape score in the context of tool construction (also shown in Figure \ref{fig:shape_scoring}):
\begin{itemize}
    \item \textbf{Independent shape scoring}: This approach separately scores each object used for the tool construction. The final shape score is then computed as a product of their independent shape scores;
    \item \textbf{Joint shape scoring}: This approach scores the \textit{combination} of the different objects in terms of the shape appropriateness of their overall configuration.
\end{itemize}
Note that, we only use joint shape scoring for tool substitution, since substitution does not involve attaching different objects together, instead the overall shape of the object is scored. We now describe each scoring method. 

\subsubsection{Independent shape scoring ($\phi_{shape}$)}
Since independent shape scoring evaluates tools on a per-part basis, we consider tools to have action parts and grasp parts\footnote{This covers the vast majority of tools \cite{myers2015affordance, abelha2017learning}}. We then train independent neural networks that can learn correspondence between the shape and function of specific tool parts. Hence, we train separate networks for the tools' action parts, and for a supporting function: ``\textit{Handle}'', which refers to the tools' grasp part.

We represent the shape of the input object point clouds using Ensemble of Shape Functions (ESF) \cite{wohlkinger2011ensemble} which is a 640-D vector. Each neural network takes the ESF feature for an object as input, and outputs a binary label indicating whether the object is suitable for the function. For more information on training the networks, we refer the reader to \cite{nair2019autonomous}. 

For the score prediction, given an action and a tuple of candidate objects $T_i$, we can compute a shape score for $T_i$ using the trained networks. Ordering of the objects within the tuple indicates correspondence to action or grasp parts. Let $\mathcal{K}$ denote the set of objects in $T_i$ that are candidates for the action parts of the final tool, and let $T_i - \mathcal{K}$ be the set of candidate grasp parts. Then the shape score $\phi_{shape}(T_i)$ is computed by using the trained networks as follows:
\begin{align*}
    \phi_{shape}(T_i) = \prod_{c_j \in \mathcal{K}}p(action|c_j) \prod_{c_j \in T_i-\mathcal{K}}p(handle|c_j)
\end{align*}
Where, $p$ is the prediction confidence of the corresponding network. Thus, we combine prediction confidences for all action parts and grasp parts. For example, if the specified action is ``hit'' and $T_i$ consists of two objects $(c_1, c_2)$, then $\phi_{shape}(T_i) = p(hit|c_1)*p(handle|c_2)$. 

\subsubsection{Joint shape scoring ($\phi'_{shape}$)}
For joint shape scoring, our goal is to learn the correspondence between the full tool shape and functionality, rather than a part-based approach. Here, we train independent dual neural networks on full tool point clouds corresponding to different actions. 

As before, we represent the input point clouds using ESF features. Each dual neural network takes as input the ESF feature for an object (or object combination), and outputs a binary label indicating whether the input is suitable for a particular function. The training procedure is similar to material scoring (additional details in \cite{shrivatsav2020}), and is used to learn an embedding space, $\mathcal{E}^p_{action}$, representative of the positive training examples (i.e., canonical tools for performing the action, obtained from tool databases such as ToolWeb \cite{abelha2016model}).  

In the case of tool substitution, each candidate object point cloud $c_i$, is passed as input to the trained model and the shape score is computed as follows:
\begin{align*}
    \phi_{shape}'(c_i) = \sigma (\mathbf{w}^T |\mathcal{E}^p_{action} - f(c_i)|^2 + \beta)
\end{align*}
Where $f$ denotes the output of the final hidden layer of the dual neural network. This score represents the similarity between the ESF feature of the input and the embedding.

For tool construction, given an input set of objects $T_i$, our joint shape scoring approach begins by aligning the components in $T_i$ in a configuration consistent with prototypical tools used for the specified action. In order to retrieve this configuration, we sample one random tool from the ToolWeb dataset used for training the shape scoring model, corresponding to the specified action. Further, we use Principal Component Analysis (PCA) to orient the object point clouds in $T_i$ with respect to the example tool. The aligned point cloud is then passed as input to the dual network to compute a shape score as above. Figure \ref{fig:shape_scoring} shows an example of the aligned point cloud. The joint shape scoring method effectively treats constructions as substitutes.

\begin{algorithm}[t]
		\SetKwInOut{Input}{input}\SetKwInOut{Output}{output}
		
		\Input{candidate tool parts $T_i$, attachment type $t_{att}$}
		\Output{$\phi_{att}(T_i)$, $A_{close}(T_i)$}
		\BlankLine
				
		$\phi_{att}(T_i) = 0,$
		$A_{close}(T_i) = []$

		$T_i' = Align(T_i)$ 
	          
		$P = ComputeIntersections(T_i')$ 
		
		\tcp{Compute attachment points based on attachment type}

		\uIf{$t_{att} = `pierce'$}{
		    \uIf{$isPierceable(T_i)$}{
		        $A^{T_i} = P$
		        
		        $\alpha = 0.5$
		    }
		    \Else{
		        $A^{T_i} = \varnothing$
		    }
		}
		\uElseIf{$t_{att} = `grasp'$}{
		    $A^{T_i} = GraspSample(T_i)$
		    
		    $\alpha = 0$
		}
		\uElseIf{$t_{att} = `magnetic'$}{
		    \tcp{Predefined magnet location}
		    $A^{T_i} = userInput(T_i)$ 
		    
		    $\alpha = 0$
		}
		\Else{
		    $A^{T_i} = \varnothing$
		}
		
		\uIf{$A^{T_i} \neq \varnothing$}{
			\ForEach{$t_i \in T_i', c_k \in t_i$}{
				$A^{T_i}(c_k) = ClosestAttachment(P, c_k, A^{T_i})$
						
				$\phi_{att}(T_i) \stackrel{+}{=} \|P, A^{T_i}(c_k)\|$ \tcp{Dist to P}
				
				$A_{close}(T_i).append(A^{T_i}(c_k))$
			}
		}
		\Else{
		    $\phi_{att}(T_i) = \infty$
		    
			\Return $\phi_{att}(T_i), P$
		}
		
		$\phi_{att}(T_i) \stackrel{+}{=} \alpha$ \tcp{Add cost}
		
		$\gamma = -max(\phi_{att}(T_i))$ \tcp{normalizer}
		
		\Return $\phi_{att}(T_i)/\gamma, A_{close}(T_i)$
		
		\caption{Attachment Fit}
\end{algorithm}

\subsection{Attachment Scoring ($\phi_{att}$)}

Given an action and a set of objects, we seek to predict whether the objects can be attached to perform the specified action. Hence, attachment scoring is specific to tool construction only. The degree to which the objects facilitate the desired attachment is indicated by the attachment score. In order to attach the objects, we consider three attachment types, namely, \textit{pierce attachment} (piercing one object with another, e.g., foam pierced with a screwdriver), \textit{grasp attachment} (grasping one object with another, e.g., a coin grasped with pliers), and \textit{magnetic attachment} (attaching objects via magnets on them). The attachment scoring algorithm is shown in Algorithm 2 ($AttachmentFit()$). Attachment scoring begins by aligning the components of the candidate tool $T_i$ in a configuration consistent with prototypical tools used for the specified action ($Align()$, line 2). This is similar to the aligned point cloud generation process followed in our joint shape scoring approach. This results in a set of alignments $T_i'$. We then approximate the intersections of the point clouds in each alignment by calculating the centroid of closest points between the point clouds ($ComputeIntersection()$, line 3). The resultant set of centroids, $P$, is the candidate list of attachments we want to make, i.e., the target attachment locations. The attachment score $\phi_{att}(T_i)$ is then computed as the Euclidean proximity of the target attachment locations, $P$, and the closest attachments facilitated by the candidate objects (denoted as $A_{close}(T_i)$), depending on the attachment type $t_{att}$. This is computed for each object $c_j$, in each alignment $t_i \in T_i'$ ($ClosestAttachment()$, lines 21-25). The resulting score, $\phi_{att}$, is normalized (by $\gamma$) (Line 31, Algorithm 2). The negative normalizer ranks lower $\phi_{att}$ as better. If a object $c_j \in T_i$ is known to have no attachment points, $\phi_{att}(T_i) = \infty$, since the objects cannot be attached to construct the tool. Thus, given the set of closest attachment points $A^{T_i}(c_j)$ on the objects $c_j \in T_i$, the attachment score is computed as (Algorithm 2, line 23): 
\begin{align*}
    \phi_{att}(T_i) = 
    \begin{cases}
      \alpha + \sum\limits_{c_j \in T_i} \left\Vert P - A^{T_i}(c_j)\right\Vert, & \text{if attachable} \\
      \infty, & \text{otherwise}
    \end{cases}
\end{align*}

The term $\alpha$ denotes a fixed cost of attaching the objects and varies for each attachment type, depending on whether some attachments are costlier than others. E.g., piercing an object may damage it, and a high cost associated with piercing can encourage other alternatives where available. Thus, the set of attachment points $A^{T_i}$ is required to compute $\phi_{att}$. In the case of pierce and grasp attachments, we assume that the capabilities of the acting tool is known ($t_{att}$ is known). That is, objects with pierce capability (screwdrivers and sharp pointed objects), and objects with grasp capability (pliers, tongs) are known a-priori. However, these can be identified using existing affordance learning approaches \cite{AffordanceNet18}. Below, we describe how the attachments are computed for each attachment type (Lines 4-19, Algorithm 2).

\subsubsection{Pierce Attachment}
Similar to material reasoning, we use the SCiO sensor to reason about material pierceability. We train a neural network to output a binary label indicating pierceability of the input spectral reading. We assume homogeneity of materials, i.e., if an object is pierceable, it is uniformly pierceable throughout the object. 

For our model, we use a neural network with a single hidden layer of 256 units and a binary output layer. We used the Adam optimizer with ReLU activation layer, and a sigmoid in the final layer. To train our model, we used the same dataset used for material reasoning, namely, SMM50, with spectrometer readings for five classes of materials: plastic, wood, metal, paper and foam. Of these classes, we consider paper and foam objects to be pierceable and for each, we provide the pierceability labels. For each material class, 12 different objects were used with 50 samples collected per object from different locations of the object. This results in a total of 600 spectrometer readings per class. 

To determine the attachment score during tool construction for the input $T_i$, the SCiO sensor is used to scan the objects and the corresponding spectral reading is passed to the classifier. The attachment score $\phi_{att}(T_i)$ is then computed based on the classifier label. If the output label is zero (Algorithm 2, line 5, $isPierceable(T_i) = 0$), $A^{T_i} = \varnothing$ since pierce attachment is not possible. If pierceable, $A^{T_i} = P$, assuming homogeneity of material properties allowing the objects to be configured at the desired location, and $\alpha = 0.5$ indicating a fixed cost of performing the pierce attachment.

\subsubsection{Grasp Attachment}
Grasp attachment is defined as using one object to grasp/hold another object to extend the robot's reach (e.g., grasping a bowl with pliers). We model the grasping tool (pliers or tongs) as an extended robot gripper, allowing the use of existing robot grasp sampling approaches \cite{ten2017grasp, levine2018learning, zech2016grasp}, for computing locations where the tool can grasp objects. In particular, we use the approach discussed by \cite{ten2017grasp}, that outputs a set of grasp locations, given the input parameters reflecting the attributes of the pliers/tongs used for grasping. We cluster the grasp locations (using Euclidean metric) to identify unique grasps. As described in their work, without any additional training, the geometry-based grasp sampling approach achieves an accuracy of 73\%. To further improve accuracy, it is possible to train an object-specific model to identify valid grasps. A key challenge with using a pre-trained model is the need to re-train it for every newly encountered pliers/tongs with differing parameters, which can be inefficient in terms of computational resources. Hence, we use the geometry-based grasp sampling approach without any object-specific refinement. 

To compute attachment score for the input $T_i$, grasps are sampled for the objects (Line 12, $GraspSample(T_i)$) using the existing grasp sampling algorithm\footnote{Implementation at https://github.com/atenpas/gpg based on \cite{ten2017grasp}}. Once sampled, the resultant grasp locations are returned as potential attachment points $A^{T_i}$. The grasp locations are used to compute $\phi_{att}$ based on their Euclidean proximity to $P$. We set $\alpha = 0$ since there is no explicit cost associated with performing grasp attachments. 

\subsubsection{Magnetic Attachment}
We assume the locations of magnets to be provided or predefined, i.e., $A^{T_i}$ is known, and we compute $\phi_{att}$ based on their Euclidean proximity to $P$. We set $\alpha = 0$ since there is no explicit cost associated with performing magnetic attachments. If magents are absent, $\phi_{att} = \infty$. However, as described in \cite{nair2019toolconstr}, it is also possible to perform magnetic attachments via exploration if they are not predefined. This process uses the desired target locations $P$ to explore attachments of the objects. If magnets are present proximal to $P$, enabling the desired configuration, then the objects are attached during the exploration process.

\begin{figure*}[t]
	\centering
	\includegraphics[width=0.95\textwidth]{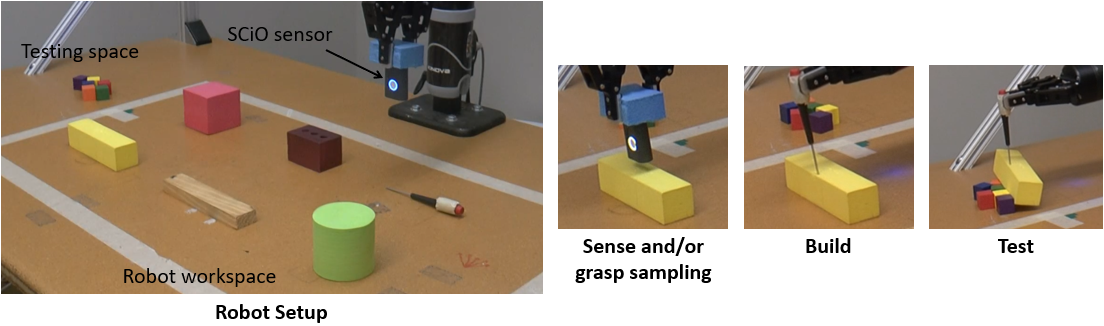}
	\captionsetup{width=\linewidth}
	\caption{The robot setup and steps involved in a typical tool construction cycle. In the case of pierce attachment, the robot uses the SCiO sensor to sense material properties, and in case of grasp attachment, the robot samples valid grasps for the object. The robot then builds the tool and tests it by performing the action with the tool \cite{nair2019autonomous}.}
	\label{fig:construction_example}
\end{figure*}

\subsection{Final Score Computation}
Given the shape, material and attachment scores, we compute the final scores for tool substitution and construction. 

\subsubsection{Tool Substitution}
The final score for tool substitutes is computed as a weighted sum of the shape and material scores. We empirically determined uniform weights of $\lambda_1 = 1$ and $\lambda_2 = 1$ to work best. Our final score for tool substitutes, $\Phi^{subs}$, is computed as follows. Note that we use the joint shape scoring method to compute shape scores for tool substitutes:
\[\Phi^{subs}(c_i) = \phi_{shape}'(c_i) + \phi_{mat}(c_i)\]

Each $c_i \in C$ denotes a candidate object, which is a potential substitute tool.

\subsubsection{Tool Construction}
For the tool constructions, the final score is computed as a weighted sum of the shape, material and attachment scores. Similar to substitution, we found uniform weights of $\lambda_1 = 1, \lambda_2 = 1$, and $\lambda_3 = 1$, to work best for tool constructions. Our final score, $\Phi^{cons}$, is computed as:
\[\Phi^{cons}(T_i) = \phi_{shape}(T_i) + \phi_{mat}(T_i) + \phi_{att}(T_i)\]

Each $T_i \in T$ denotes a permutation of the candidate objects used for tool construction. Note that either the independent or joint shape scoring approach can be used towards the final score computation for tool construction. If joint shape scoring is used, the final score is computed as $\Phi^{subs}(T_i) + \phi_{att}(T_i)$. The final score can optionally be used to generate a ranking of tool constructions. The robot can then iterate through the ranking until a successful construction is found \cite{nair2019autonomous}.

\subsection{Arbitration of Tool Substitution and Tool Construction}
Arbitration combines tool substitution and tool construction within our pipeline, and in this section we present different arbitration strategies for deciding between the two. We formulate the problem as follows:

\textit{``Given an action, and a set C of n candidate objects, how can we arbitrate between tool substitution and tool construction for accomplishing the specified action?''}


\smallskip Inspired by existing research in behavioral robotics, each strategy (substitution or construction) is associated with a value function, $\Psi$, that dictates the strategy chosen at a given instant \cite{velayudhan2017sloth, arkin2003ethological, likhachev2000robotic}. The value functions in our work, account for the overall fitness of the substitutes and constructions for performing the specified action. We generate a combined ranking of the strategies (highest to lowest value) that the robot iterates through, validating each strategy until a solution is found. Our set of states $S = C \cup T$, represents the union of the set of all individual objects $c_i$, and the set of all permutations of $m$ objects $T_i$, for tool construction. We now introduce three different value functions for arbitration, that uses the final scores computed in the previous section.


First, we present a \textbf{rule-based approach} that assigns a fixed value to constructions, as follows:
\begin{align*}
    \Psi_{rule}(s_i) = 
    \begin{cases}
      10, & \text{if} \ |s_i| = 1, \Phi^{subs}(s_i) > 1.0 \\
      0, & \text{if} \ |s_i| > 1, \Phi^{cons}(s_i) > 1.0 \\
      -\infty, & \text{otherwise}
    \end{cases}
\end{align*}
Where, $|s_i|$ denotes the cardinality of $s_i \in S$, to indicate whether a single object is being evaluated (substitute, $c_i$) or a combination of objects (construction, $T_i$). This approach prefers substitutions over constructions, provided the substitutions have a higher score than a threshold. We empirically set our threshold to 1.0. A fixed value is also assigned to constructions that exceed the threshold in terms of the construction objective. 

Second, we present a \textbf{direct comparison} approach that compares objectives, and assigns values to states in $S$ as:
\begin{align*}
    \Psi_{obj}(s_i) = 
    \begin{cases}
      \Phi^{subs}(s_i), & \text{if} \ |s_i| = 1 \\
      \Phi^{cons}(s_i), & \text{if} \ |s_i| > 1
    \end{cases}
\end{align*}
Note that the tool construction objective $\Phi^{cons}(s_i)$ automatically assigns a cost associated with attachments, namely the attachment score $\phi_{att}$, and penalizes constructions over substitutions. Here, tool construction uses the independent shape scoring approach. 

Third, we present a \textbf{substitution-based} approach that uses joint shape scoring for tool constructions, in effect treating the constructions as substitute objects. Hence, the values for states in $S$ are assigned as follows:
\begin{align*}
    \Psi_{subs}(s_i) = 
    \begin{cases}
      \Phi^{subs}(s_i), & \text{if} \ |s_i| = 1 \\
      \Phi^{subs}(s_i) + \phi_{att}(s_i), & \text{if} \ |s_i| > 1
    \end{cases}
\end{align*}
Here, the final attachment score is added to account for the cost of attachment for tool constructions. This enables the tool constructions and substitutions to be compared directly in terms of the shape scoring objective. 

In the following sections, we evaluate each component of our Tool Macgyvering pipeline, namely, tool construction, tool substitution, and arbitration. 




\section{Tool Construction Evaluation}

\begin{figure}[t]
	\centering
	\includegraphics[width=0.49\textwidth]{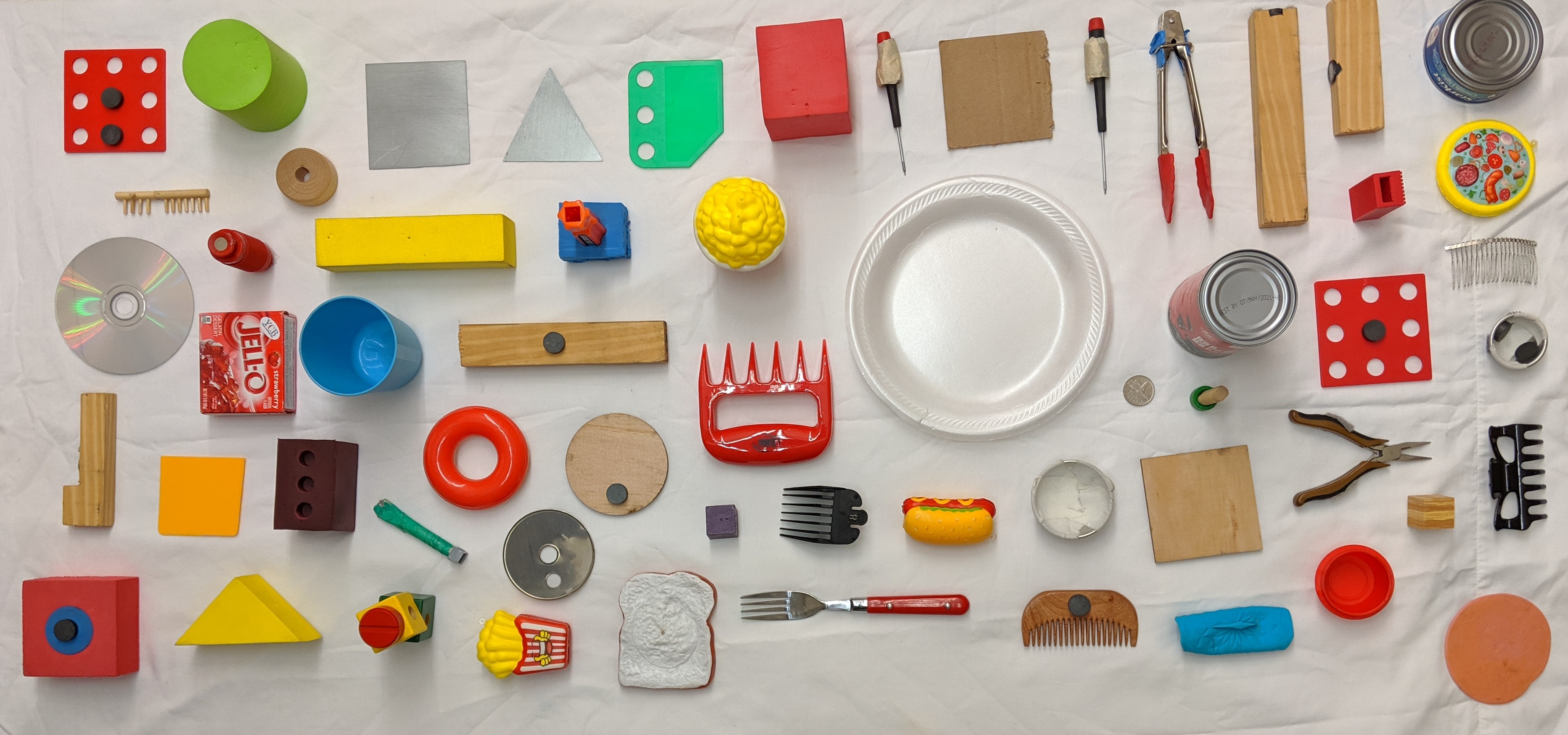}
	\captionsetup{width=\linewidth}
	\caption{The 58 objects used for experimental validation.}
	\label{fig:all_obj}
\end{figure}

\begin{table*}[t]
	\centering
	\includegraphics[width=0.94\textwidth]{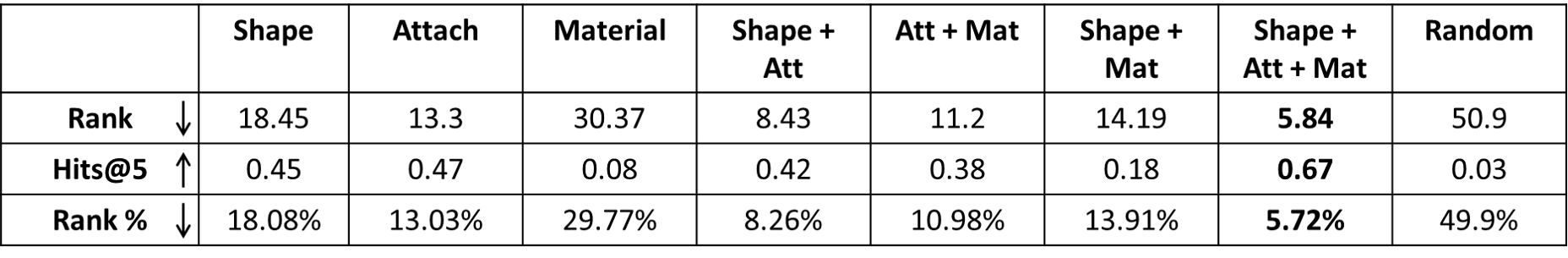}
	\captionsetup{width=\linewidth}
	\caption{Table showing results of our ablation studies. Combined shape, material and attachment reasoning (in bold) performs best. Arrows indicate whether lower or higher values are preferred, e.g., lower ranks are preferred.}
	\label{fig:final_pipeline}
\end{table*}

In this section, we describe our experimental setup and present the results specifically for our tool construction approach. We validate our approach on the construction of tools for six different actions, encoded as textual inputs: `hit', `scoop/contain', `flip', `screw', `rake' and `squeegee'. Each tool consists of two components ($m = 2$) corresponding to the action part (`hit', `scoop/contain', `flip', `screw', `rake', `squeegee') and grasp part (`handle'). The performance of our tool construction approach is evaluated in terms of the final ranking output by our algorithm. We use the final score $\Phi^{cons}$ to rank the different constructions. The tool models used to compute the desired attachment location $P$, is acquired from the ToolWeb dataset \cite{abelha2017learning}. Our experiments seek to validate two aspects of our work:
\begin{enumerate}
    \item \textit{Final tool ranking evaluation}: Performance of our tool construction approach in terms of final ranking, with ablation studies. We use the final score $\Phi^{cons}$ to rank the object constructions.
    \item \textit{Comparison to prior tool construction approaches}: Performance of our current tool construction approach against our prior work, namely, \cite{nair2019toolconstr, nair2019autonomous}\footnote{As discussed in the Related Work, we are not aware of any other prior work that demonstrates tool construction using environmental objects.}.
\end{enumerate}

For all our experiments, we use a test set consisting of 58 previously unseen candidate objects for tool construction (shown in Figure \ref{fig:all_obj}). These objects consist of metal (11/58), wood (12/58), plastic (19/58), paper (2/58) and foam (14/58) objects. Only the foam and paper objects are pierceable. Figure \ref{fig:construction_example} shows a sample experimental setup and steps involved in the robot tool construction. During tool construction, the robot begins by scanning the materials of the objects for attachment scoring, followed by ranking and construction of the tools. The robot then tests the tool by using it to perform the desired action, iterating through the ranks until a successful construction is found. To overcome manipulation and perception challenges that are beyond the scope of this work, the available objects were spaced apart and oriented to facilitate grasping. 

For the evaluation, we create 10 different sets of 10 objects (chosen from the 58) for each of the six tools, and report the average results (total $10 \times 6$ cases with 10 candidate objects per case). We create each set by choosing a random set of objects, ensuring that only one ``correct'' combination of objects exists per set. The correct combinations are determined based on human assessment of the objects.

\begin{table*}[t]
	\centering
	\includegraphics[width=0.9\textwidth]{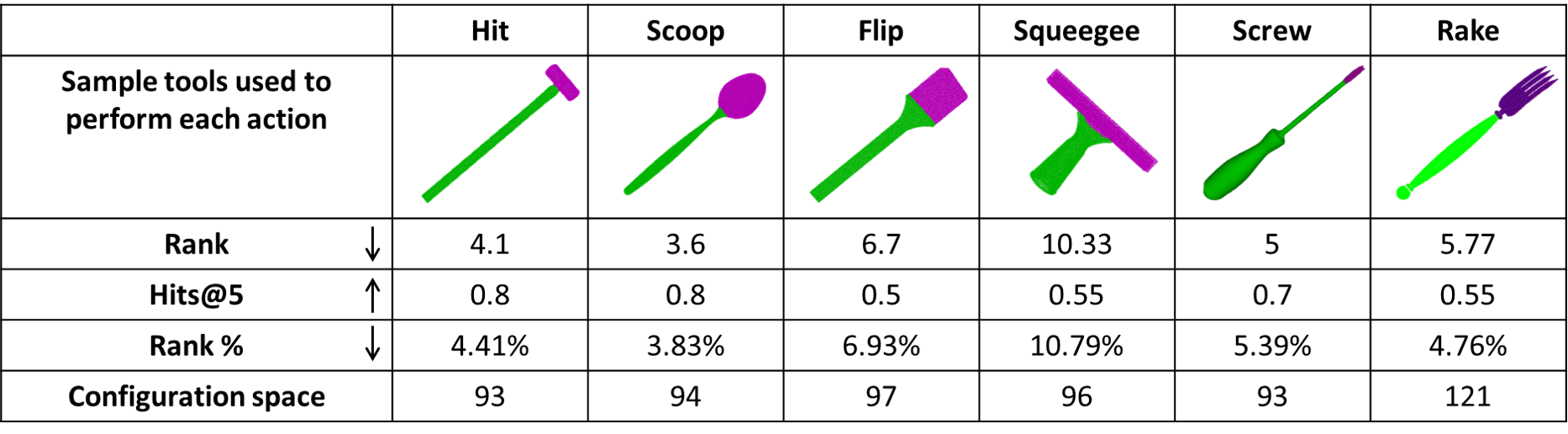}
	\captionsetup{width=\linewidth}
	\caption{Table showing tool-wise breakdown of the combined shape, material and attachment reasoning approach along with the example tools used for computing the target attachment locations.}
	\label{fig:indi_tools}
\end{table*}

\subsection{Final Tool Ranking}
\label{subsec:final_ranking}

We evaluate our overall approach in terms of the final output ranking generated on the sets of objects described in the previous section. We perform ablation studies to compare performances of shape, material and attachment reasoning for tool construction. For shape scoring, we use the independent shape scoring approach owing to its success in our previous work \cite{nair2019autonomous}.


The metrics used in this evaluation consider i) the final ranking of the correct combinations, and ii) the computation time. We would like the correct combination to be ranked as high as possible, ideally ranked at 1, indicating that it would be the first object combination the robot will attempt to construct. We report the average rank of the correct combination for each tool (average of 10 builds), the number of builds for which the correct combination was ranked within the top 5 ranks (hits@5), the average number of possible configurations of objects, and the average total computation time. The number of object configurations highlight the complexity of the state space and is also used to compute the rank\% as the fraction of rank over total configuration space.

Table \ref{fig:final_pipeline}, shows the overall performance of our approach, and Table \ref{fig:indi_tools} shows a tool-wise breakdown. From Table \ref{fig:final_pipeline}, we see that our final approach combining shape, material and attachment scoring, yields a rank of 5.84, with 67\% hits@5, and 5.72\% rank\%. Hence, we see that there is a significant benefit to combining shape, attachment and material reasoning, in terms of final ranking, rank\% and hits@5. Using only shape and attachment also performs well with a rank of 8.43 and rank\% of 8.26\%, in comparison to the other baselines. All approaches significantly outperform random ranking, which explores roughly half of the entire configuration space (with rank\% of 49.9\%).  

In Table \ref{fig:indi_tools}, we show the performance of combined shape, material and attachment reasoning for each action. Also shown are some example tools used for the computation of the target attachment locations, $P$. Overall, our approach achieved an average rank of 5.84 across all tool types. Note that the total configuration space for each tool is large (avg. $\approx 100$ configurations), indicating the complexity of the problem space, and the effectiveness of our combined reasoning approach in ranking the tool construction with a rank\% of 5\%. Thus, only a small fraction of the total configuration space is explored by our approach. We also note that the approach performed relatively worse on ``squeegee'' with a rank of 10.33, primarily because none of the available object combinations closely resemble an actual squeegee, making it a challenging problem for tool construction. Our approach achieves an average rank of 5.03 across the remaining tool types.

\textit{\textbf{Summary:}}
Combining shape, material and attachment reasoning leads to significantly improved performance for tool construction compared to prior work. 

\subsection{Comparison to tool construction approaches}

\begin{figure*}[t]
	\centering
	\includegraphics[width=0.98\textwidth]{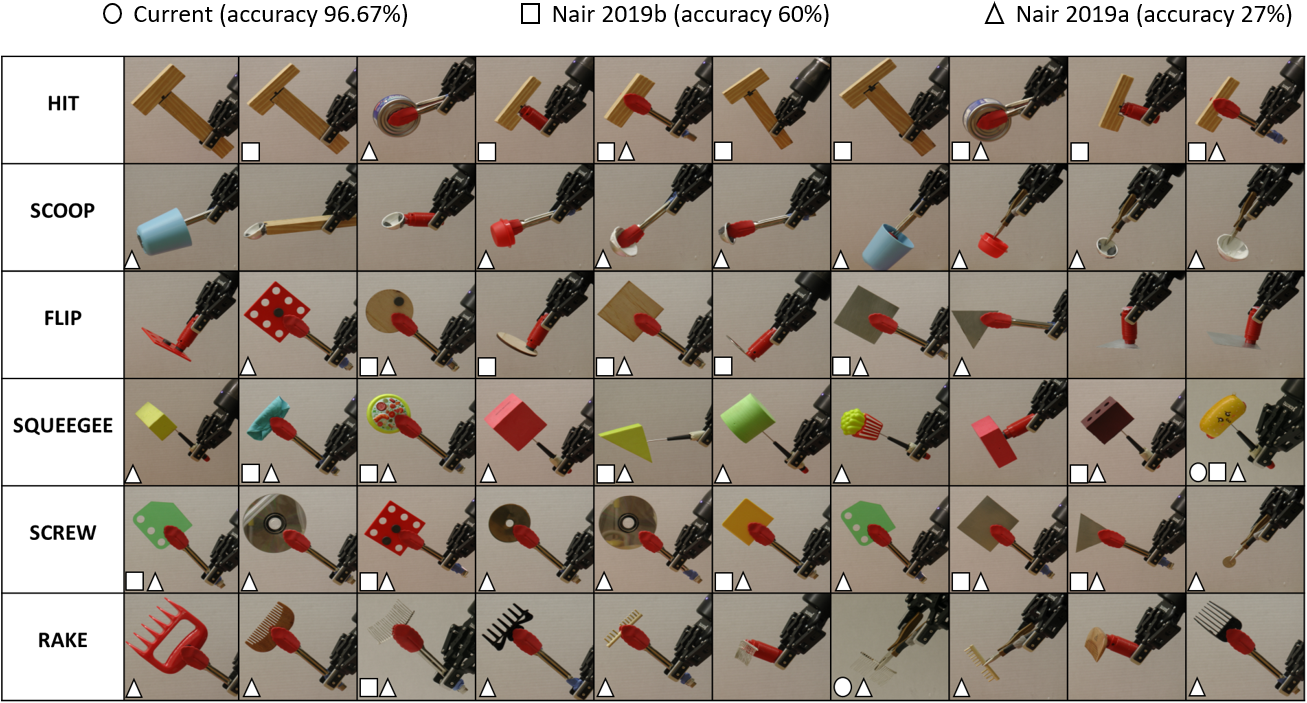}
	\captionsetup{width=\linewidth}
	\caption{Table showing a collage of the complete 60 tool constructions in our test set, constructed for six different actions. Note that a small number of experiments (8/60) led to the creation of similar tools due to the availability of objects that could be connected. A symbol on the bottom left of each image indicates that a given approach failed to find the correct construction in that case: $\circ$ : Current work, $\square$: \cite{nair2019autonomous}, and $\triangle$: \cite{nair2019toolconstr}.}
	
	\label{fig:tool_table}
\end{figure*}

We compare our final tool construction approach incorporating material reasoning, to our prior work, namely, \cite{nair2019toolconstr} and \cite{nair2019autonomous}. We use the same set of objects and evaluation metrics as the previous section, additionally adding the completion rate metric to indicate how many of the total 60 constructions, were successfully found. We mark a tool construction attempt as a failure if either, 1) the correct combination was assigned a score of $-\infty$, e.g., due to incorrect attachment/material predictions or, 2) the approach returned a tool that did not match in terms of material, e.g., hammers constructed of foam.

\begin{table}[t]
	\centering
	\includegraphics[width=0.49\textwidth]{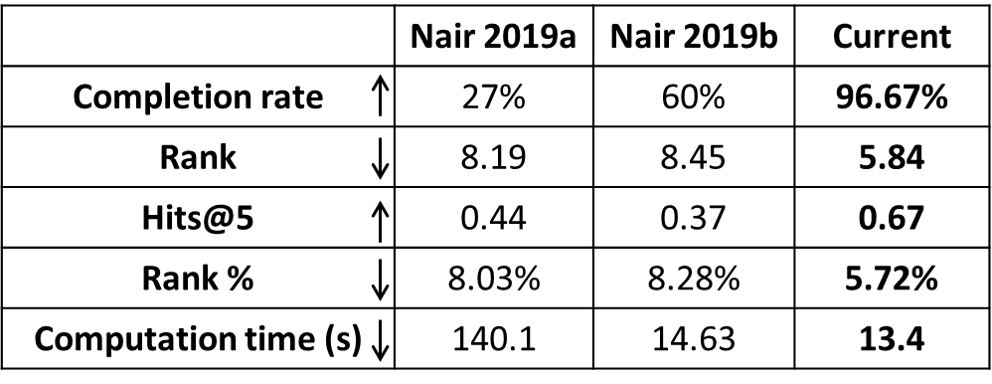}
	\captionsetup{width=\linewidth}
	\caption{Table showing performance of our current approach against previous tool construction work (\cite{nair2019toolconstr, nair2019autonomous}).} 
	\label{fig:baselines_final}
\end{table}

Our results are shown in Tables \ref{fig:baselines_final} and Figure \ref{fig:tool_table}. As shown in Table \ref{fig:baselines_final}, our current approach outperforms our prior work with a high completion rate of 96.67\%, rank of 5.84 and hits@5 of 67\%. Hence, there is an improvement in the tool construction pipeline with the introduction of material reasoning, reflected by the lower completion rates of the other approaches (27\% for \cite{nair2019toolconstr} and 60\% for \cite{nair2019autonomous}). Our approach fails at some constructions owing to incorrect pierceability and graspability predictions.

Figure \ref{fig:tool_table} shows the diversity of tool constructions output by our approach, including several interesting combinations, e.g., combining pliers and coin to create screwdriver (Construction \#10). The symbols at the lower left corner indicate \textit{failed} constructions for each approach. Note that, 91\% of the failure cases in our prior approaches were owing to incorrect materials of the constructed tools. Overall, our current approach is able to effectively reason about materials, resulting in improved quality of constructions over prior work. Additionally, our results demonstrate the capability of our approach to construct a diverse set of tools.

\textit{\textbf{Summary:}}
Incorporation of material reasoning significantly improves the performance of tool construction over prior approaches, with improved quality of constructions.

\section{Tool Substitution Evaluation}

\begin{figure}[t]
	\centering
	\includegraphics[width=0.49\textwidth]{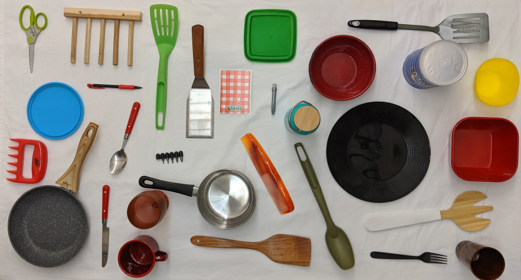}
	\captionsetup{width=\linewidth}
	\caption{The 30 objects used for evaluating tool substitution.}
	\label{fig:subs_dataset}
\end{figure}

In this section, we briefly summarize results from our prior tool substitution work \cite{shrivatsav2020} for six actions: ``Hit'', ``Cut'', ``Scoop'', ``Flip'', ``Poke'' and ``Rake'', with five material classes: Metal, wood, plastic, paper and foam. Our experiment validated the performance of combined shape and material reasoning for tool substitution on a set of partial point clouds, and spectral readings of real-world objects. The 30 objects used in our experiments are shown in Figure \ref{fig:subs_dataset}. For validation, we created six sets of 10 objects per action (total 36 sets).  Each set consisted of one ``correct'' substitute for the given action, and nine incorrect, which acts as our ground truth\footnote{The correct substitute was determined by three independent evaluators (with a Cronbach's alpha of 0.93).}. We used the final score $\Phi^{subs}$ to rank the different tool substitutes and evaluate our approach. Our metrics included hit@1, indicating the proportion of sets for which the correct tool was ranked at 1; Average Rank, which is the average rank of the correct tool across the test sets; and hits@5, indicating the number of times the correct tool was ranked within the top five ranks of our output.

Our results in Table \ref{fig:results_final} show that overall, our approach combining shape and material outperformed the other conditions, with an average ranking of 2 across all the sets. In particular, we note that combining shape and material significantly improved hit@5 (86\% vs 67\% for shape and 58\% material only), and hit@1 (53\% vs 28\% for shape and 22\% material only). All three approaches performed significantly better than random ranking of the objects (hit@1 of 5\% and hit@5 of 14\%).

\begin{table}[t]
	\centering
	\includegraphics[width=0.42\textwidth]{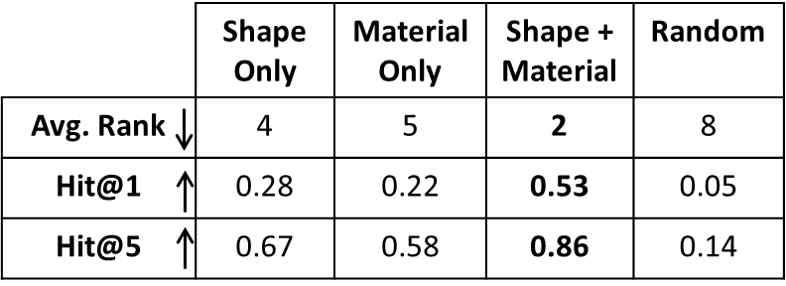}
	\captionsetup{width=\linewidth}
	\caption{Chart showing the ablation results for tool substitution. Combined shape and material scoring performs better overall (bold). Arrows indicate whether higher or lower values are preferred \cite{shrivatsav2020}.}
	\label{fig:results_final}
\end{table}

\begin{figure}[t]
	\centering
	\includegraphics[width=0.49\textwidth]{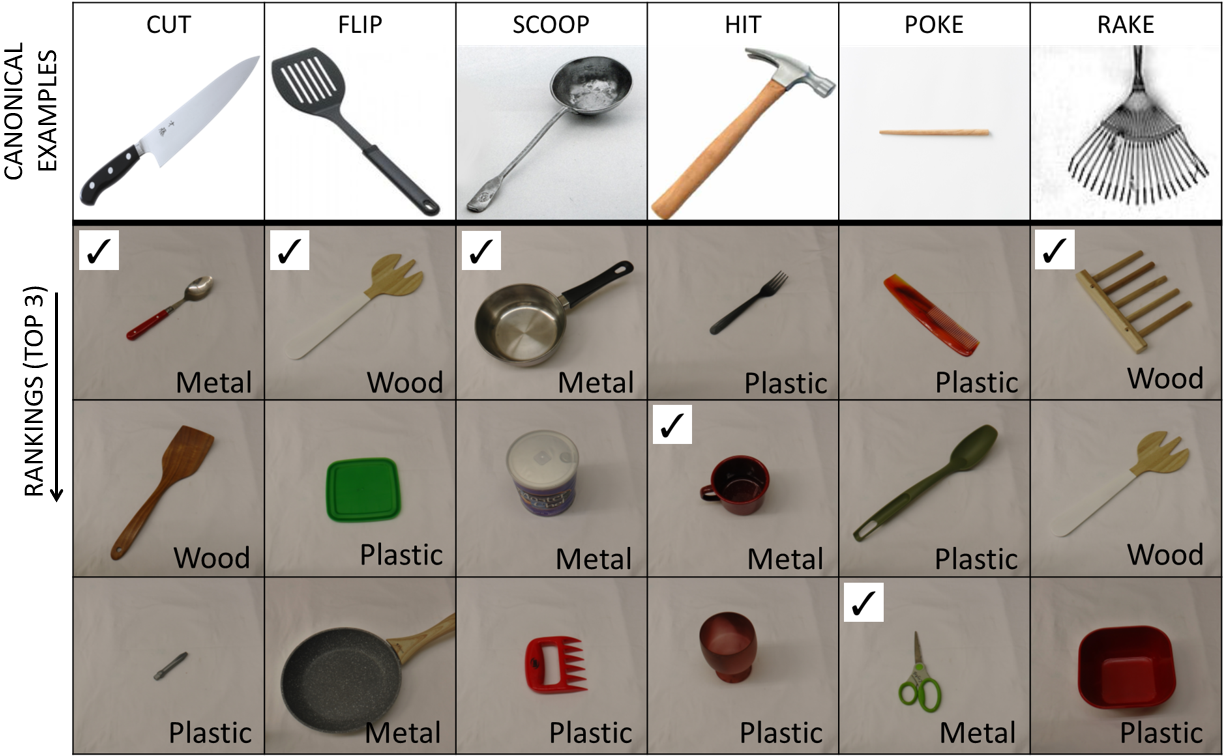}
	\captionsetup{width=\linewidth}
	\caption{First row shows examples of some canonical tools for each action. Following rows show the ranking of objects (top 3) for some of the sets. Check marks indicate the ground truths. The actual materials of the objects are also noted \cite{shrivatsav2020}.}
	\label{fig:collage}
\end{figure}

Figure \ref{fig:collage} shows some of the ranked substitutes returned by combined shape and material reasoning, for some of the test sets. The results highlight the challenges of working with partial RGBD data and material scans. For example, the (closed) metal can ranked as the \#2 substitute tool for scooping is ranked highly, because its reflective surface resulted in a point cloud that resembled a concave bowl. Further, an incorrect material prediction for the metal mug, resulted in it being ranked as \#2 substitute for hitting.

\textit{\textbf{Summary:}}
Combined shape and material reasoning leads to significantly improved performance for tool substitution, when compared to reasoning about material only or shape only.

\section{Evaluation of Arbitration Strategies}

Given the independent evaluations of tool construction and substitution presented above, we now evaluate how these two capabilities can be combined using different arbitration strategies.
As before, we validate our strategies on six different actions: `hit', `scoop', `flip', `screw', `rake' and `squeegee'. We created five different sets of objects per action for a total of 30 different cases. In each set, we included one ``correct'' substitute object, and one ``correct'' constructed object (substitution/construction pair), both of which are capable of performing the action, and the remainder of the objects were randomly chosen incorrect candidates. The ``correct'' substitutes and constructions for each pair were selected from the substitution and construction test sets used in the experiments described previously. In each case, we asked three independent evaluators (with Cronbach's alpha of 0.93), to evaluate which among the substitute/construction pair would be a better alternative for performing the specified action. For each object in the test set, the final scores were computed ($\Phi^{subs}$ or $\Phi^{cons}$), and used in the value functions for arbitration. Thus, the final ranking generated by the value functions is a combined ranking of tool substitutes and constructions. We evaluated our arbitration strategies, both in the context of the overall ranking of the ground truth, and also in terms of the specific option chosen between the two alternatives (i.e., either substitution or construction).

Our evaluation metrics include average rank, rank\%, and hits@5, as before. Additionally, we include a metric that indicates the \% times the correct option was chosen (\% correct). We compute this by evaluating whether the arbitration strategy correctly chose between the substitution/construction pair i.e., scored the ground truth option better. 

\begin{table}[t]
	\centering
	\includegraphics[width=0.47\textwidth]{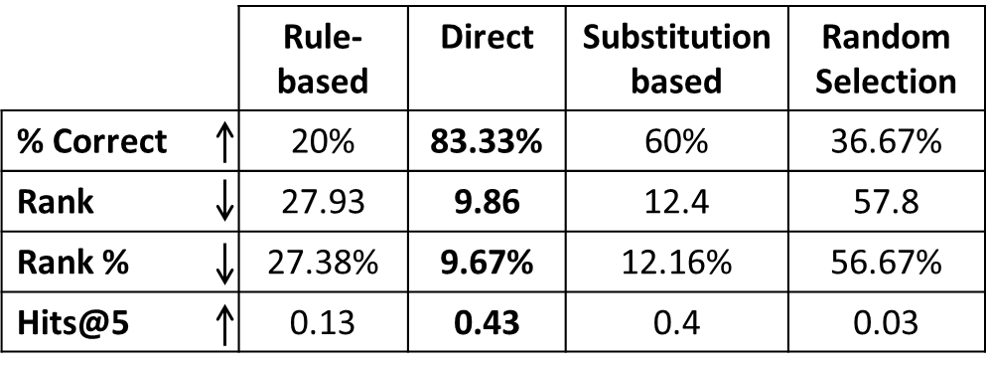}
	\captionsetup{width=\linewidth}
	\caption{Chart showing the \% number of times the correct option was chosen by each arbitration approach, along with other metrics. Bold highlights the best approach, and arrows indicate whether higher or lower values are preferred.}
	\label{fig:arb_compare}
\end{table}

Our results in Table \ref{fig:arb_compare} show that direct comparison of scores outperform the other approaches. In terms of ranking (rank, rank\% and hits@5), we note that both direct and substitution-based approaches perform comparably. However, in terms of the \% times the correct strategy was chosen, direct comparison (83.33\%) outperformed substitution-based approach (60\%). In our observations, the substitution-based approach was more likely to rank substitutes as better than constructions. However, both direct comparison and substitution-based approaches outperformed random selection (36.67\%). Another observation is regarding the inferior performance of the rule-based approach (20\%) compared to random selection, in terms of \% correct. This is because rule-based almost consistently ranked substitutes as better than constructions, which did not always conform with the ground truth labels. However, it performed better in terms of average rank, rank\% and hits@5. This is because the ground truth substitutes were ranked better consistently, resulting in a better average ranking performance.

\begin{figure}[t]
	\centering
	\includegraphics[width=0.49\textwidth]{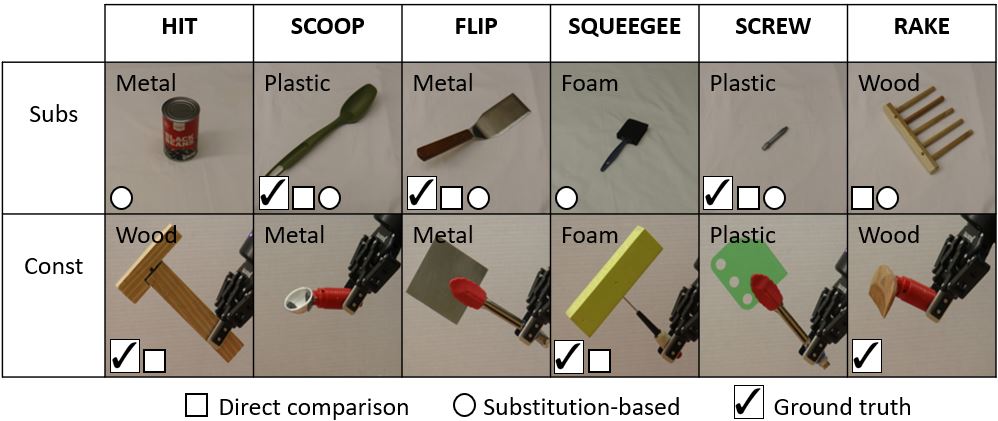}
	\captionsetup{width=\linewidth}
	\caption{The arbitration results for direct and substitution-based approach for six substitution/construction pairs. Checkmarks indicate the human evaluated ground truth. Subs $\rightarrow$ substitution, const $\rightarrow$ construction.}
	\label{fig:arb_collage}
\end{figure}

Our results in Figure \ref{fig:arb_collage} highlights some of the substitution-construction pairs in our test set, along with the selections of the direct and substitution-based strategies. As shown in the table, substitution-based approach was more inclined towards selecting the substitute tools over the constructions. Overall, the direct comparison approach conformed more to the ground truth assessments made by the human evaluators. 

\textit{\textbf{Summary:}} Direct comparison of the scores outperformed other baseline approaches in arbitrating between construction and substitution. Secondly, the best design choices for our final Tool Macgyvering framework involve using independent shape scoring (combined with material and attachments) for tool construction; joint shape scoring with material reasoning for tool substitution; and direct comparison for arbitration.

\section{Discussion and Future Work}
In this work, we presented a novel Tool Macgyvering framework that combined tool substitution and tool construction using arbitration, to output macgyvered solutions for performing an action. We extended our prior work on tool construction by incorporating material reasoning, resulting in significantly improved performance and quality of output constructions. Our approach effectively discovered 96.67\% of working object combinations (as opposed to 27\% and 60\% in prior work), while exploring only a small percentage of the total configuration space (5.72\%). We also introduced arbitration strategies for deciding between tool substitution and construction for performing an action. Our arbitration strategy involving direct comparison of scores correctly selected between substitution and construction for 83.33\% of the test cases, outperforming the other approaches. In summary, the key findings of this work are as follows: 
\begin{enumerate}
    \item Combining material reasoning with shape and attachment reasoning significantly improves quality of output constructions, with a superior performance over previous tool construction approaches in terms of completion rate (96.67\% completion);
    \item Combined material, shape and attachment reasoning enables the efficient construction of a wide range of tools as shown in Figure \ref{fig:tool_table}; 
    \item Arbitration by direct comparison correctly selected between substitution and construction with an accuracy of 83.33\%, and performed better than other arbitration strategies; 
    \item The best performing design for our final Tool Macgyvering framework includes: a) tool construction utilizing \textit{independent shape scoring}, \textit{material scoring}, and \textit{attachment scoring}, b) tool substitution utilizing \textit{joint shape scoring} and \textit{material scoring}, combined with c) \textit{direct comparison} for arbitration.
\end{enumerate}


In future work, a number of changes can be made to further improve the performance of the system.  For example, we observed cases in which shape scoring produced incorrect ranking because the RGBD sensor captured only a partial point cloud of an object. Future work can address such problems through active perception.  Additionally, our future work will address a key limitation of our current approach that the number of objects utilized for constructing a tool equals number of tool parts i.e, there is a one-to-one correspondence between candidate objects and tool parts. Finally, additional physical attributes, such as mass and density, can be incorporated into the reasoning framework to further improve performance.

In terms of arbitration, a key limitation of our existing approach is that it only considers the physical attributes of the objects. However, other factors such as effort, risk, and the task constraints can influence the decision. In our future work, we will expand our arbitration strategies to consider a wider range of factors within a multi-objective function.

\ifCLASSOPTIONcaptionsoff
  \newpage
\fi



%

\section*{Acknowledgments}
This work is supported in part by NSF IIS 1564080 and ONR N000141612835.

\bibliographystyle{./IEEEtran}
\bibliography{references}

%








\end{document}